\documentclass[10pt,twocolumn,letterpaper]{article}

\usepackage[pagenumbers]{cvpr} 

\definecolor{cvprblue}{rgb}{0.21,0.49,0.74}
\usepackage[pagebackref,breaklinks,colorlinks,allcolors=cvprblue]{hyperref}
\usepackage{times}  
\usepackage{helvet}  
\usepackage{courier}  
\usepackage[hyphens]{url}  
\usepackage{graphicx} 
\usepackage{natbib}  
\usepackage{caption} 
\usepackage{xcolor}
\usepackage{algorithm}
\usepackage{algorithmic}
\usepackage{verbatim}
\usepackage{array}
\usepackage{multirow}
\usepackage{amssymb}
\usepackage{amsmath}
\usepackage{threeparttable}

\makeatletter
\newcommand*\bigcdot{\mathpalette\bigcdot@{.5}}
\newcommand*\bigcdot@[2]{\mathbin{\vcenter{\hbox{\scalebox{#2}{$\m@th#1\bullet$}}}}}
\makeatother

\title{Unity in Diversity: Multi-expert Knowledge Confrontation and Collaboration for Generalizable Vehicle Re-identification}

\author{Zhenyu Kuang\footnotemark[1]\\
Xiamen University\\
\and
Hongyang Zhang\footnotemark[1]\\
Xiamen University\\
\and
Mang Ye\\
Wuhan University\\
\and
Bin Yang\\
Wuhan University\\
\and
Yinhao Liu\\
Xiamen University\\
\and
Yue Huang\\
Xiamen University\\
\and
Xinghao Ding\\
Xiamen University\\
\and
Huafeng Li\\
Kunmimg University of Science and Technology
}

\usepackage{bibentry}

\begin{document}
\maketitle
\begin{abstract}
Generalizable vehicle re-identification (ReID) seeks to develop models that can adapt to unknown target domains without the need for additional fine-tuning or retraining. Previous works have mainly focused on extracting domain-invariant features by aligning data distributions between source domains. However, interfered by the inherent domain-related redundancy in the source images, solely relying on common features is insufficient for accurately capturing the complementary features with lower occurrence probability and smaller energy. To solve this unique problem, we propose a two-stage Multi-expert Knowledge Confrontation and Collaboration (MiKeCoCo) method, which fully leverages the high-level semantics of Contrastive Language-Image Pretraining (CLIP) to obtain a diversified prompt set and achieve complementary feature representations. Specifically, this paper first designs a Spectrum-based Transformation for Redundancy Elimination and Augmentation Module (STREAM) through simple image preprocessing to obtain two types of image inputs for the training process. Since STREAM eliminates domain-related redundancy in source images, it enables the model to pay closer attention to the detailed prompt set that is crucial for distinguishing fine-grained vehicles. This learned prompt set related to the vehicle identity is then utilized to guide the comprehensive representation learning of complementary features for final knowledge fusion and identity recognition. Inspired by the unity principle, MiKeCoCo integrates the diverse evaluation ways of experts to ensure the accuracy and consistency of ReID. Extensive experimental results demonstrate that our method achieves state-of-the-art performance.
\end{abstract}
\section{Introduction}

Vehicle re-identification aims to locate and identify all instances of the same vehicle within video surveillance networks, which is one of the critical issues in the field of computer vision. In recent years, significant progress has been made in ReID due to the rapid development of deep learning and the spectacular debut of data-centric large models \cite{r88, r14, r20, r21, r18, r89, r90, li2024AAAI}. Supervised vehicle ReID methods are based on the assumption of independent and identically distributed data, which assumes that the training and test set have the same data distribution. However, in diverse and dynamic real-world application scenarios, the ideal experimental setup is often disrupted due to significant differences in data distributions between the source and target domains. As a result, well-trained models face a significant decline in performance because of the domain shift problem \cite{r91}, which arises when the data distribution during training differs from that during testing.

\begin{figure}[t!]
\centering
\includegraphics[width=3.2in,height=1.6in]{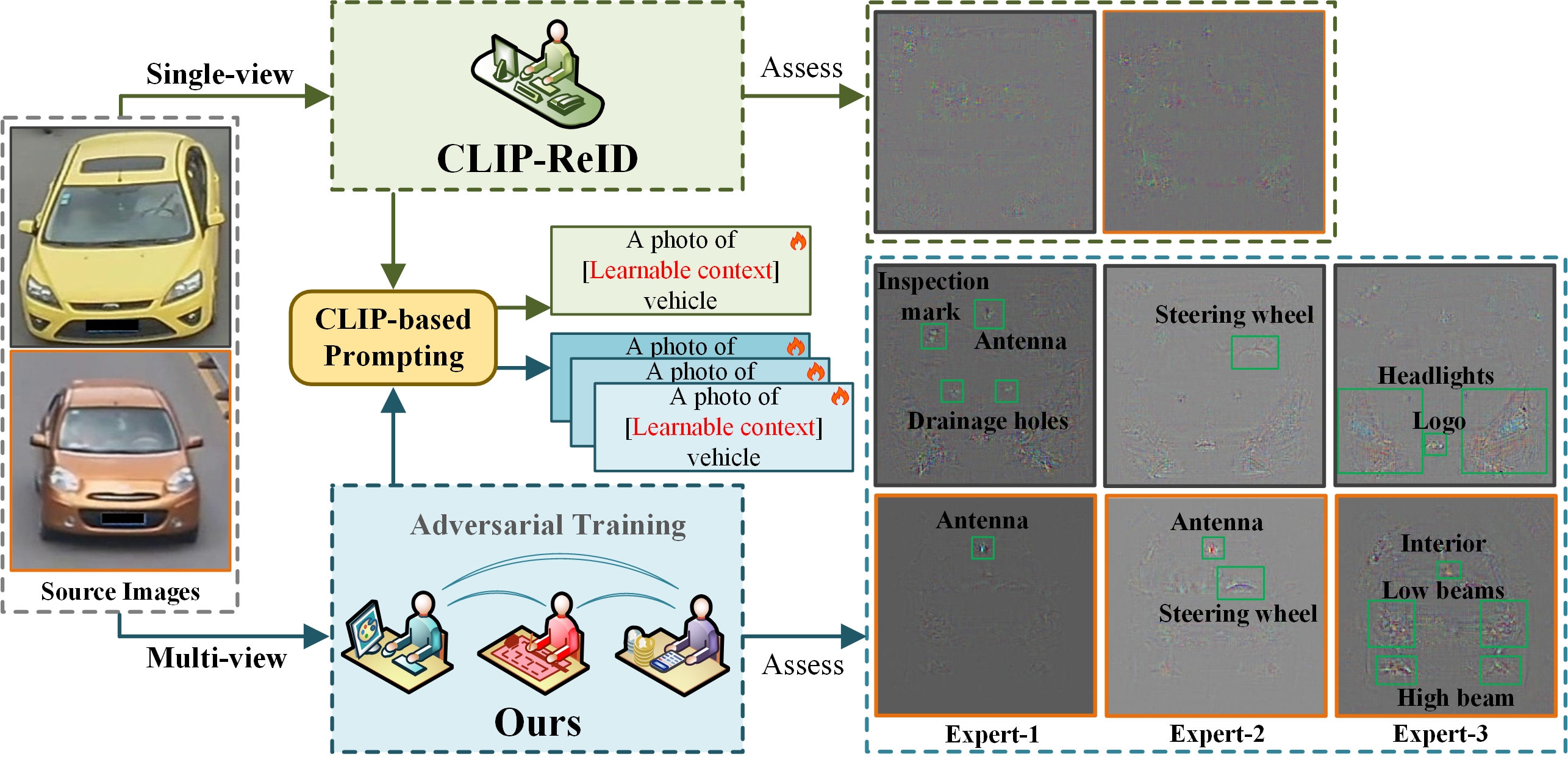}
\caption{Backpropagation-based visualizations \cite{r95} of saliency map of CLIP-ReID and our method. CLIP-ReID method of adapting CLIP to single-view prompt learning has not yet eliminated the domain-related redundancy of source images, which restricts its diversified feature extraction. Through domain-related redundancy elimination and multi-view adversarial training, our method effectively integrates the complementary features of different high-level semantics that are of concern to three different experts.}
\label{fig:2}
\end{figure}

Domain adaptation (DA) methods \cite{r92, r93} are considered one of the primary ways to addressing the domain shift problem, which enable models to adaptively align the data distributions between the source and target domains. But DA methods rely on vast amounts of data from both the source and target domains for effective alignment. Besides, when the data from the target domain is heavily noisy, the effectiveness of domain adaptation may be significantly reduced. Its practical application is constrained due to the demand for target domain datasets. To address challenges in practical applications, domain generalization (DG) methods \cite{r7, r8, r23, r24, r65} further train on single or multiple source domain datasets to obtain domain-invariant feature representations, which can be applied to unknown target domains. While many domain generalization methods have been proposed for person ReID, the problem has not yet received sufficient attention in the context of vehicle ReID.

In contrast to person ReID, vehicle ReID faces the unique problem of highly similar appearances \cite{r71} among different vehicles of the same type and color. In this case, the limited and discriminative features are challenging for deep neural networks to extract effectively. Nevertheless, individual features of vehicle instances still exhibit significant local variability, such as scratches, interior details, annual inspection mark, and antennas. Even within the same type of identifying feature, there may be differences in position, shape, and size across different vehicle instances. If a single feature vector is used to uniformly measure all individuals, the differences in feature occurrence probability and energy contribution could cause a few features to dominate the entire feature learning process. As a result, features with lower occurrence probability and smaller energy may not be effectively learned, leading to inaccurate recognition of vehicle instances containing these features in DG tasks. On the other hand, in DG vehicle ReID tasks, images are often accompanied by domain-related components that are not related to vehicle identification. Therefore, eliminating domain-related redundancy in vehicle images can enable the model pay more attention to the complementary features with lower occurrence probability and smaller energy.

\begin{figure}[t!]
	\centering
        \includegraphics[width=1\linewidth]
       {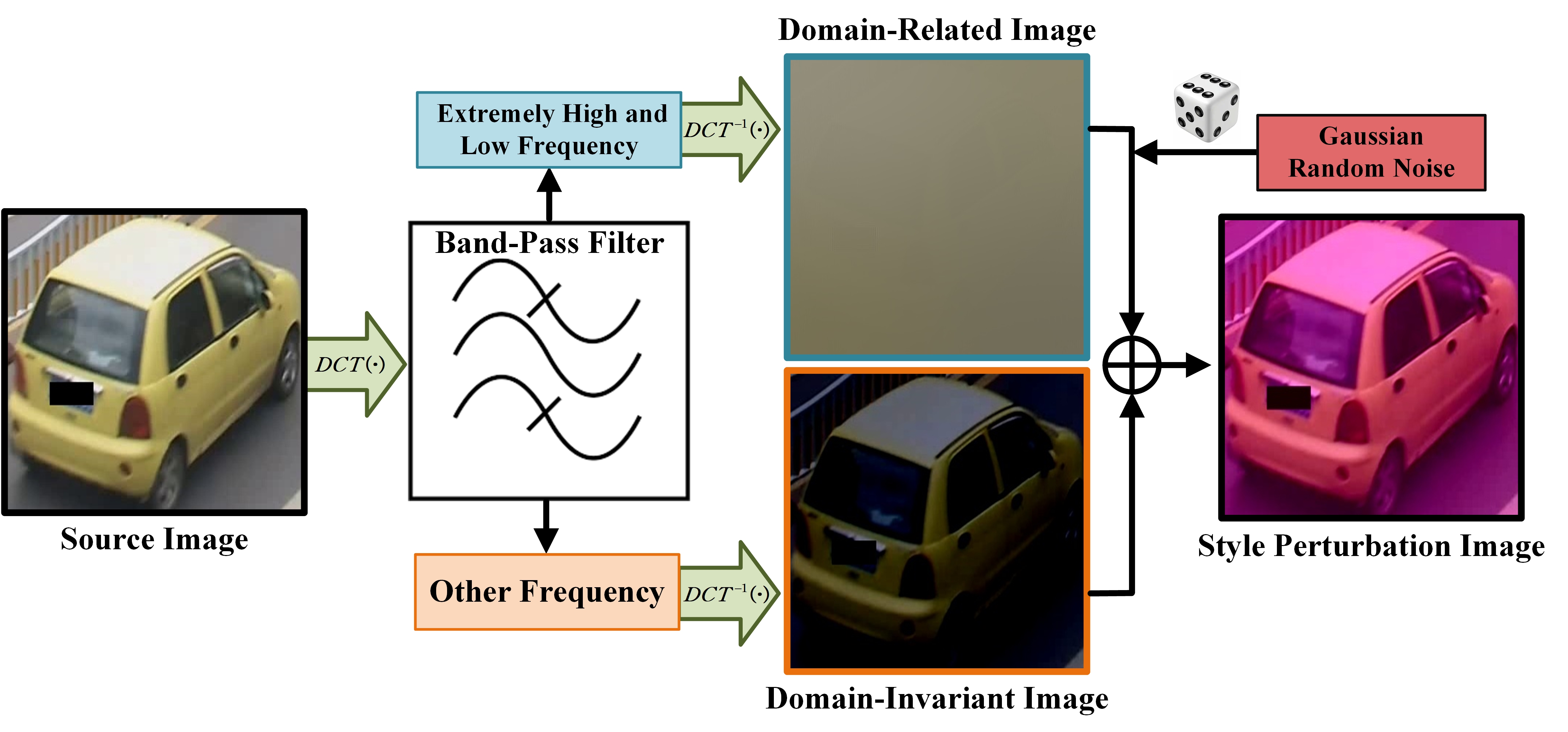}
	\caption{Schematic diagram of STREAM. Inspired by the conclusion of reference \cite{yang2020fda, s3}, the extremely high and low frequency component of an image contains primary domain-related information. $DCT(\bigcdot)$ and $DCT^{-1}(\bigcdot)$ represent the forward and inverse discrete cosine transforms, respectively. Domain-invariant image and style perturbation image are used respectively for the input images of the two-stage training of MiKeCoCo.}
	\label{causal}
\end{figure}

While most existing methods \cite{r61, r62, r64, r65, r70} focus on ID-related feature representations at the feature level by designing novel model architectures and objective functions, they may overlook the inherent domain-related redundancy in source images. This redundant information directly participates in the training process of deep neural networks, still limiting the models' ability to extract generalizable features. Another class of image augmentation methods \cite{s4, r96} based on generative adversarial networks attempts to increase data diversity through image style transformation and data augmentation. But the problems of data noise and redundancy remain unresolved. This not only results in the generated images containing a large amount of task-irrelevant redundant information but also leads to unnecessary consumption of computational resources in reconstructing ID-unrelated content, thereby reducing training efficiency. 
Recently, the most relevant work to ours is CLIP-ReID \cite{r18}. CLIP-ReID demonstrates exceptional recognition performance in the object ReID tasks, it has yet to address the interference caused by redundant information in source domain images during prompt learning. The training way with a single text prompt and domain-related redundancy in images still prevents the model from fully mining comprehensive features. The visualized results in Fig.\ref{fig:2} show the difference between CLIP-ReID and our method.

\begin{figure}[t!]
\centering
\includegraphics[width=1\linewidth]{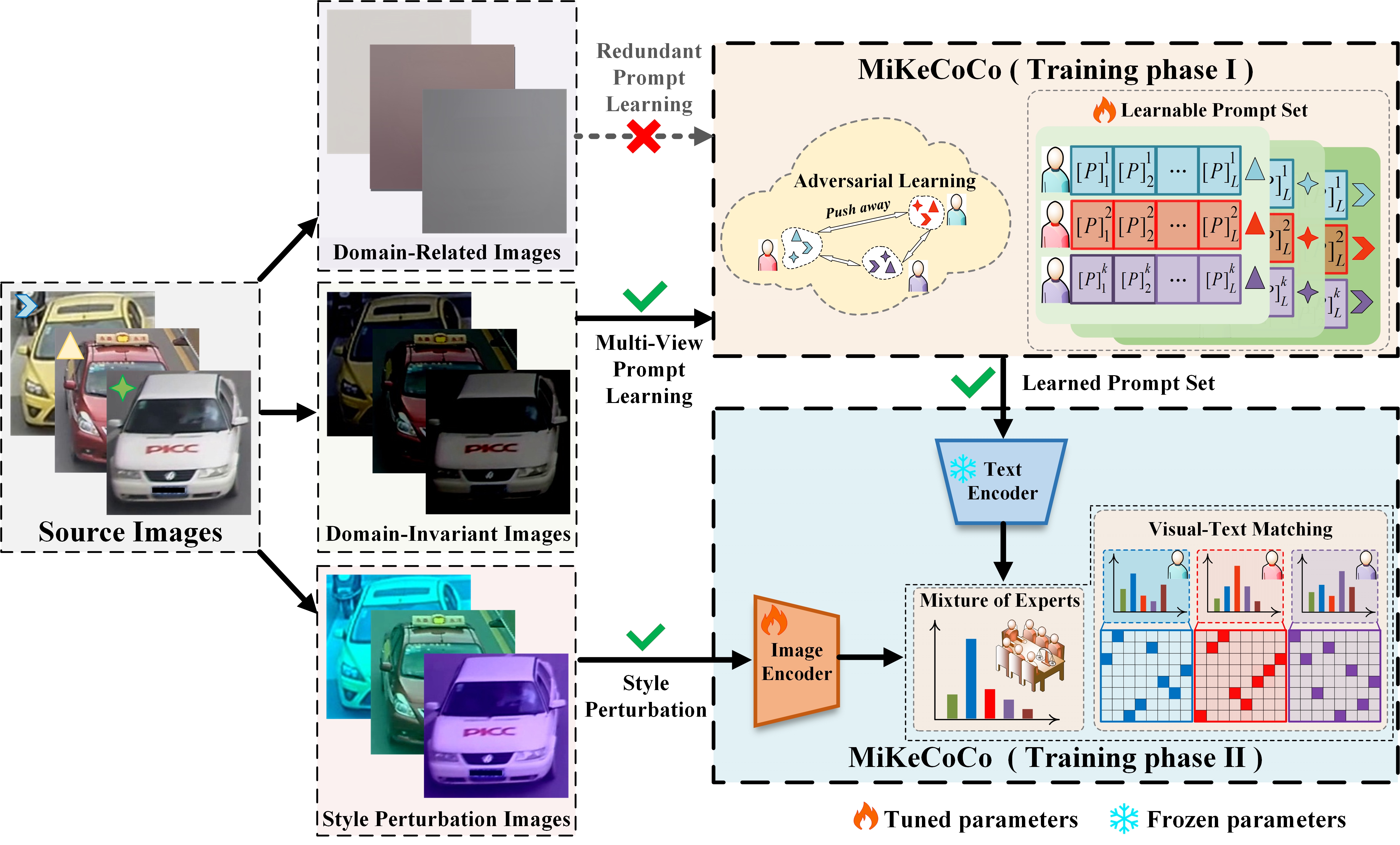}
\caption{STREAM cleverly provides the image components necessary for each of the two training stages. Since the redundant components of the source image are blocked before the first stage, domain-invariant images could be used for multi-view prompt learning inspired by adversarial training to obtain learnable prompt set related to the vehicle identity. This learned prompt set and style perturbation images are combined to achieve robust and discriminative feature representation of image encoder for the second stage. $L$ indicates the length of the learnable prompt set. }
\label{morecasual}
\end{figure}

To solve the above problems, we propose a two-stage MiKeCoCo method, which fully leverages the high-level semantics of CLIP to obtain a diversified prompt set and achieve complementary feature representations. This paper first proposes utilizing simple STREAM to eliminate domain-related redundancy from source images and to achieve data augmentation at the image level, as shown in Fig.\ref{causal} and \ref{morecasual}. To alleviate the interference caused by the domain-related factors of the source images, domain-invariant images (DII) are introduced to acquire the diversified prompt set related to vehicle identity by multi-expert adversarial learning in the first training phase. The learned prompts are less affected by domain-related factors such as illumination and background changes. And style perturbation images (SPI) can increase the diversity of source domains to enhance feature representation of image encoder of CLIP. In the second training phase, we combine all complementary features highlighted by all experts for final knowledge fusion to determine the identity of each vehicle. Ultimately, we integrate the complementary features observed from multi-expert perspectives to highlight the effectiveness of comprehensive features.

In summary, the contributions of this paper can be summarized as follows:
\begin{itemize}
\item[$\bullet$] We propose a multi-expert knowledge confrontation and collaboration framework based on the CLIP architecture. This framework facilitates the learning of multi-view complementary representations and effectively integrates visual features with the prompt set learned by CLIP.
\item[$\bullet$] We propose a spectrum-based transformation for redundancy elimination and augmentation module to eliminate domain-related redundancy in source images, enabling the model to more easily capture the intrinsic characteristics associated with core concepts of vehicle.
\item[$\bullet$] Experimental results show that our method achieves state-of-the-art performance on some public vehicle datasets.
\end{itemize}

\section{Related work}
\subsection{Domain Generalization for ReID}
In recent years, domain generalization for person re-identification has received increasing attention. 
To effectively alleviate the domain shift problem, IS-GAN \cite{r96} based on image augmentation utilizes identity-shuffling technique used by generative adversarial network and identification labels to factorize identity features without requiring any auxiliary information. Another approach to address this issue is based on the domain-invariant feature representation methods. SNR~\cite{r61} proposes a style normalization and restitution module to encourage the separation of identity-relevant features and identity-irrelevant features. The graph sampler \cite{r8} aims to establish a nearest neighbor graph for all classes at the beginning of each epoch and iteratively refine the sampling strategy. 

The above works aim to learn robust and discriminative features, but overlook the potential interference caused by domain-related redundancy inherent in the source domain images. In this work, we shift our focus towards eliminating domain-related redundancy in the image, thereby reducing unnecessary feature associations between the vehicle and the environment. It enables the model to pay closer attention to the detailed cues that may be crucial for distinguishing fine-grained vehicles.

\subsection{Visual-Language Learning}
With the rise of visual-language models \cite{r34, r37, r81}, CLIP models \cite{r16, r31, r35} have been widely applied to downstream visual tasks due to their powerful generalization capabilities and have achieved notable success. Recently, prompt learning based on CLIP method has received wide attention. CoOp \cite{r43} learns optimizable prompts for visual concepts in images to adapt pre-trained vision-language models without requiring any text annotations. Inspired by CoOp, CLIP-ReID \cite{r18}, which employs a two-stage training strategy, demonstrates exceptional recognition performance. To combine the advantages of coarse-grained attributes and implicit learnable prompts, MP-ReID \cite{r81} introduces multi-prompts learning with cross-modal alignment using integrated generative models like ChatGPT and VQA. 

Nevertheless, the above methods mainly focus on modeling the relationship between the input image and the learnable prompts. These methods have not yet addressed the interference caused by redundant components in the input images for prompt learning. As a result, it leads to the redundant prompt learning and limits the generalization performance of the models. Our method not only eliminates redundant components through simple image preprocessing, but also employs a multi-expert evaluation mechanism to derive a comprehensive representation by emphasizing the diverse
complementary features.

\begin{figure*}[t!]
	\centering
	\includegraphics[width=\textwidth]{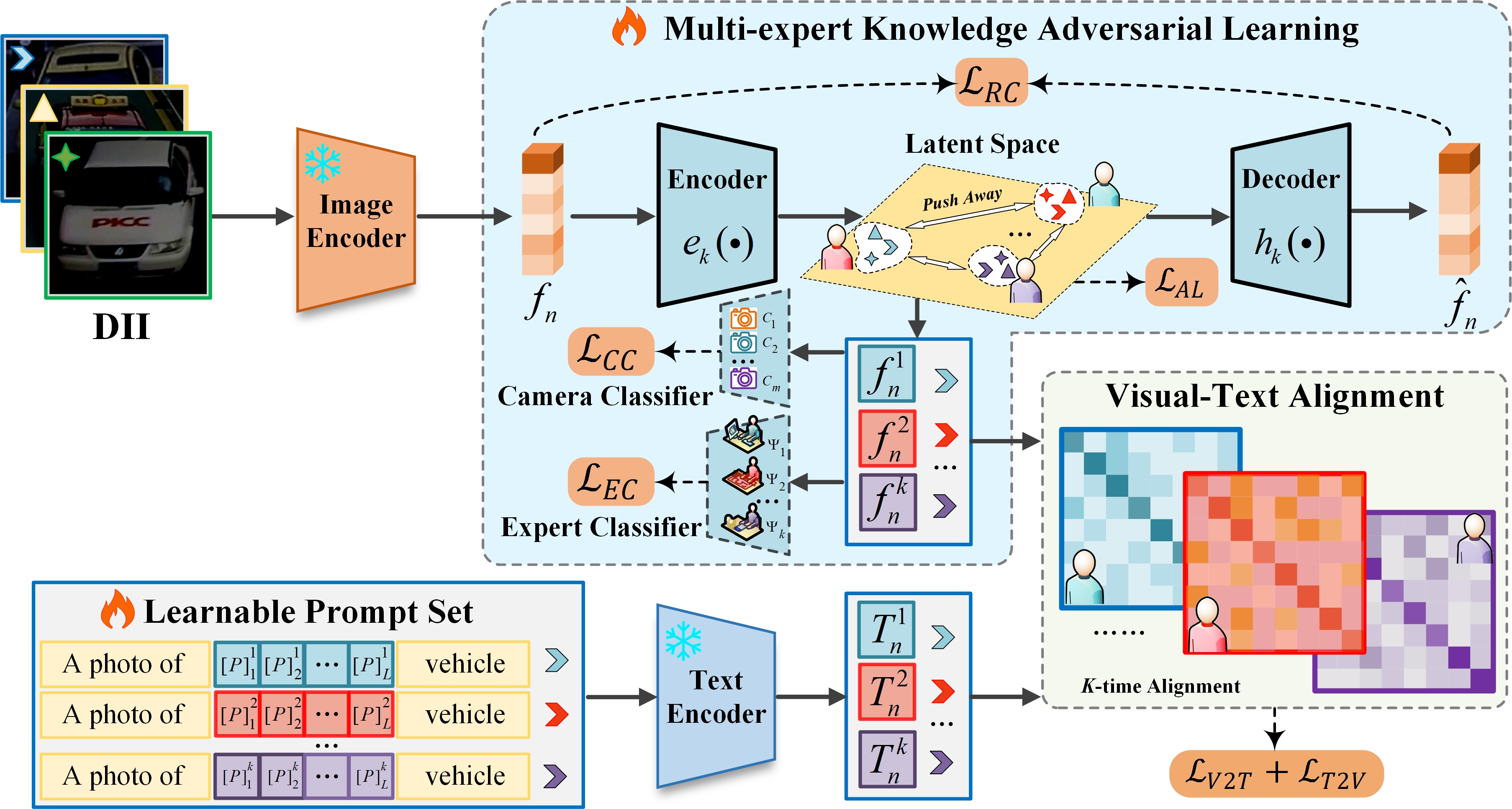}
	\caption{Overview of the first training stage of MiKeCoCo. Since DII does not contain domain-related redundancy, it can enable the model to pay closer attention to the complementary features that are crucial for distinguishing fine-grained vehicles. In this stage, the image and text encoder parameters are fixed, and the goal is to obtain a diversified prompt set by updating the MEKA module parameters.}
	\label{fig:3}
\end{figure*}

\section{Proposed method}
\subsection{Problem Statement and Overview}
\begin{figure*}[ht!]
	\centering
	\includegraphics[width=\textwidth]{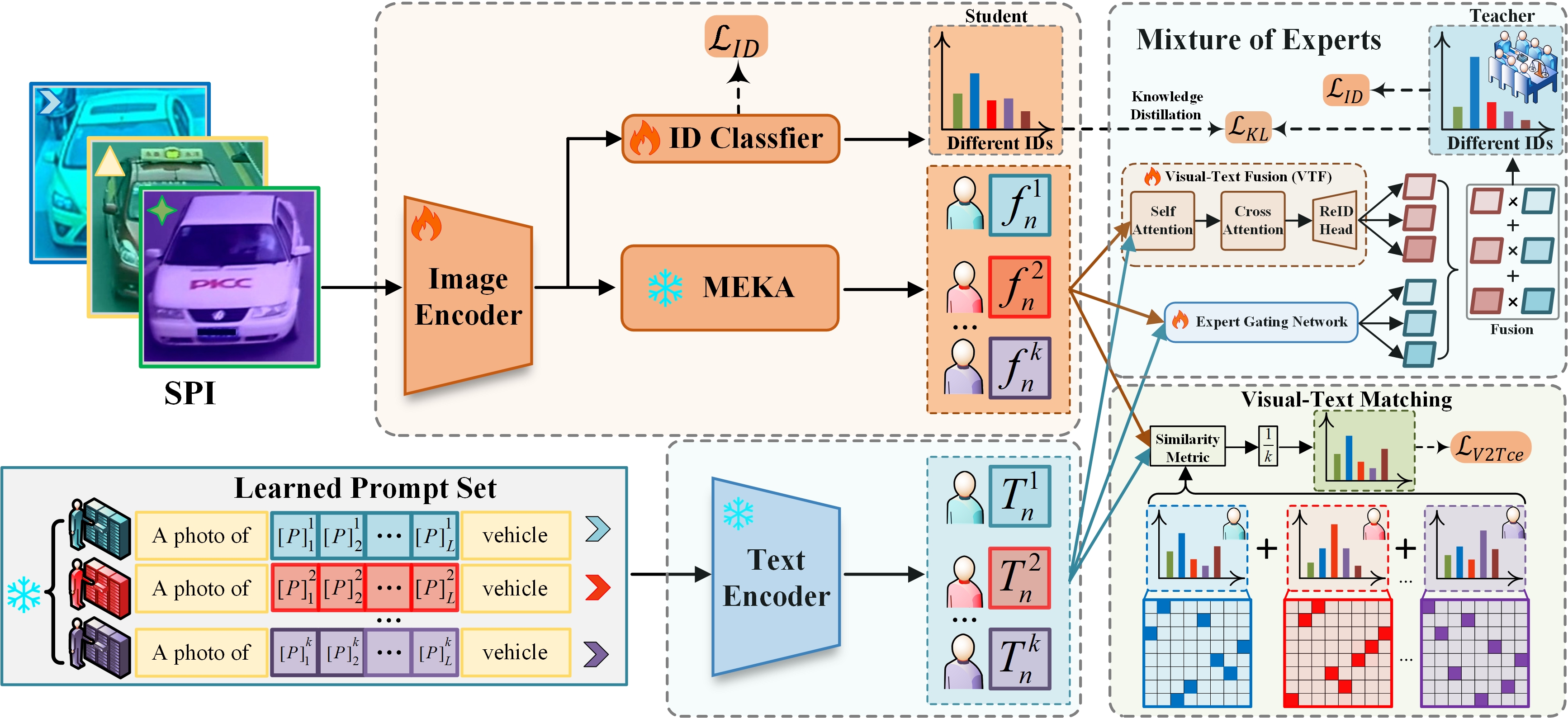}
	\caption{Overview of the second training stage of MiKeCoCo. Since SPI enhances the diversity of source domain dataset, it enables the image encoder of CLIP to extract more robust feature representations. In this stage, while multiple experts employ different assessment ways to verify the same vehicle, their common goal is to confirm the vehicle's true identity. By combining their expertise, their collective decision can enable the model to ensure the accuracy and consistency of the evaluation results.}
	\label{fig:4}
\end{figure*}

In this section, we focus on studying a generalized CLIP-based model named MiKeCoCo to tackle vehicle ReID in direct cross-domain scenarios. To eliminate image redundancy and enhance data diversity, we first design a STREAM to obtain DII and SPI from the source images, as shown in Fig. \ref{causal}. DII and SPI will be used as input images for the two-stage training of MiKeCoCo, respectively. The framework of MiKeCoCo is shown in Fig. \ref{fig:3} and Fig. \ref{fig:4}, incorporating STREAM and multi-expert perspectives to effectively acquire diverse semantic knowledge from the CLIP model for comprehensive feature extraction. Fig. \ref{fig:3} shows the process of obtaining the multi-expert prompt set in the first training stage. Fig. \ref{fig:4} illustrates the training of the image encoder guided by multi-expert knowledge fusion. We simultaneously proposes a dual-drive framework to address the domain generalization problems in vehicle re-identification, which includes a data-driven strategy based on image redundancy elimination and a model-driven approach based on multi-expert perspectives.

We have access to a labeled source dataset $\mathcal{S} = \{ X_{s}, Y_{s}^{id}, Y_{s}^{cam} \}$ with $n$ images of $N_{id}$ vehicles, where each instance is associated with a camera label and the source camera set $Y_{s}^{cam} = \{ y_{cam}^{1}, y_{cam}^{2}, ..., y_{cam}^{n} \}$. The unlabeled target domain dataset $\mathcal{T} = \{ X_{t} \}$ consists of $N_{t}$ samples.


\subsection{Spectrum-based Transformation for Redundancy Elimination and Augmentation Module}
To eliminate image redundancy and enhance data diversity, DII and SPI obtained through STREAM are used as input images for the two-stage training of MiKeCoCo, respectively. The detailed process is shown in Fig.\ref{causal}. Since DII does not contain domain-related redundant components, it enables MiKeCoCo to closely focus on the diversified prompt set related to vehicle identity. In contrast, to improve the robustness of CLIP's image encoder, SPI is introduced to increase the diversity of source images by the randomizing frequency components according to a Gaussian distribution. 

In this module, the source images $X_{s}$ can be divided into causal part $X_{cas}$ and non-causal part $X_{non}$. We design a band-pass filter $\mathcal{M}(r)$ to separate causal and non-causal factors in the frequency domain. Noises with Gaussian distribution $\mathcal{N}(0,1)$ are added into the non-causal part $X_{non}$ to obtain the augmented non-causal factors $R_G(X_{non})=X_{non}\bigcdot\big(1+\mathcal{N}(0,1)\big)$. Then, we regroup the causal and non-causal parts to get the style perturbation images $\hat{X}$ from the input images $X_{s}$ by adopting $DCT^{-1} (\bigcdot)$. The style perturbation images $\hat{X}$ are as below:

\begin{equation}\label{formula7}
	\begin{aligned}
	     \hat{X}=DCT^{-1}\Big(R_{G}\big(\mathcal{M}(r) \bigcdot DCT(X_{s})\big)+ \\
	     \big(1-\mathcal{M}(r)\big) \bigcdot DCT(X_{s})\Big),
	\end{aligned}
\end{equation}
and the domain-invariant images $\hat{X}_{cas}$ are as below:
\begin{equation}\label{formula8}
	\begin{aligned}
		\hat{X}_{cas}=DCT^{-1}\Big(\big(1-\mathcal{M}(r)\big) \bigcdot DCT(X_{s})\Big).
	\end{aligned}
\end{equation}

This reorganization process can be viewed as a pre-processing operation of the input image, which aims to separate ID-related part $\hat{X}_{cas}$ and domain-related part $X_{non}$. $\hat{X}_{cas}$ is introduced in the first training stage as DII to construct the learnable prompt sets associated with pure vehicle identity in the first stage of training. The reconstructed images $\hat{X}$ are used as style perturbation images in the second training phase. The learned prompt set with high-level semantics is then used together with SPI to compel the feature extractor of CLIP for the comprehensive features. In addition, more detailed settings for STREAM are presented in the supplementary materials.

\subsection{\textbf{Multi-expert Knowledge Adversarial Learning}}
To obtain a learnable prompt set, multi-expert knowledge adversarial learning (MEKA) maps visual features to diverse latent spaces through multiple autoencoders to explore complementary features. Motivated by the researches in multi-view learning \cite{chen2023learnable}, we fully leverage multi-view knowledge from CLIP to identify each vehicle image through adversarial learning. While different experts have unique evaluation perspectives by adversarial training, their ultimate goal is to reach the same conclusion, which is to verify the ID the vehicle image belongs to. Therefore, integrating all complementary features of multi-expert perspectives can extract richer semantic information.

The detailed process of MEKA is shown in Fig.\ref{fig:3}. Adversarial learning unfolds within diverse latent spaces of the image encoder of CLIP, comprising a expert classifier (EC), a camera classifier (CC), and multiple autoencoder networks. We employ $\mathit{K}$ experts within MiKeCoCo method and can harness the image encoder of CLIP to efficiently extract salient features $f_{n}$ from the $n$-th sample within the source domain dataset. And $f_{n}$ can be mapped to the same latent feature space by different encoders of autoencoder network, which can be represented as $\left \{f_{n}^{1},f_{n}^{2},...,f_{n}^{k} \right \}$. The reconstructed features $\hat{f_{n}} = \left \{\hat{f_{n}^{1}},\hat{f_{n}^{2}},...,\hat{f_{n}^{k}}  \right \} $ are ultimately obtained from the subsequent decoders of autoencoder network. In this autoencoder network, the model follows traditional autoencoder setting which consists of several encoders $e_{k}(\bigcdot)$ and decoders $h_{k}(\bigcdot)$, where $k\in 1,2,...,K$. To guarantee the reconstructed visual features contain full semantic information, we utilize reconstruction loss $\mathcal{L}_{RC}$ with mean-square error $MSE(\bigcdot)$ loss to constrain the distance between the input $n$-$th$ visual feature $f_{n}$ and the reconstructed features $\hat{f_{n}}$. The formulation of $\mathcal{L}_{RC}$ is described as below:
\begin{equation}\label{formula9}
	\begin{aligned}
		\mathcal{L}_{RC}= \frac{1}{\mathit{K}}\sum_{k=1}^{K}MSE(f_{n}, \hat{f_{n}^k}).
	\end{aligned}
\end{equation}

The encoder $e_{k}(\bigcdot)$ is composed of a single layer perceptron and aims to map the visual features into the multi-expert discriminator. The output of $e_{k}(\bigcdot)$ is fed into an expert discriminator $\mathit{D}_{k}$ with $\mathit{K}$ perspectives. The expert perspective label for the $i$-$th$ instance is denoted as $y_{per}^{i}$. Cross-entropy loss is employed to train the expert perspective discriminators and the formulation of $\mathcal{L}_{EC}$ is defined as below:
\begin{equation}\label{formula10}
	\begin{aligned}
     	\mathcal{L}_{EC}=-\frac{1}{N_{s}}\sum_{i=1}^{N_{s}}\sum_{k=1}^{K}y_{per}^{i} \bigcdot log\bigg(p\Big(D_{k}\big(e_{k}(f_{i}^{k})\big)\Big)\bigg),
	\end{aligned}
\end{equation}
where $N_{s}$ denotes the number of camera instances in a training batch from the source domain. Meanwhile, we propose multi-expert adversarial learning loss $\mathcal{L}_{AL}$ to ensure the autoencoder network to map visual features into more diverse complementary features.
Thus, adversarial learning loss is introduced to build multi-expert perspectives for capturing complementary features:
\begin{equation}\label{formula11}
	\begin{aligned}
		\mathcal{L}_{AL} = -\frac{\sum_{i}^{K}\sum_{j,i\ne j}^{K}\left \|e_{i}(f_{n})-e_{j}(f_{n})  \right \|^{2}}{K^{2}-K}.
	\end{aligned}
\end{equation}

While domain-invariant images no longer contain domain-related information through the STREAM, the accompanying camera label $Y_{s}^{cam}$ can still serve as supervised information and provide crucial references for prompt learning regarding viewpoint changes and the direction the vehicle is facing. Therefore, the camera classifier $\mathit{D}_{g}(\bigcdot)$ is utilized to identify the camera-specific features and the camera classification loss $\mathcal{L}_{CC}$ are described as below:
\begin{equation}\label{formula12}
	\begin{aligned}
		\mathcal{L}_{CC} = -\sum_{n=1}^{N}y_{cam}^{n}\bigcdot log\bigg(p\Big(D_{g}\big(e_{g}(f_{n})\big)\Big)\bigg),
	\end{aligned}
\end{equation}
where $\mathit{e}_{g}(\bigcdot)$ is a learnable mapping. The overall loss of MEKA with trade-off parameters $\lambda_{1},\lambda_{2},\lambda_{3}$ is described as below:
\begin{equation}\label{formula13}
	\begin{aligned}
		\mathcal{L}_{MEKA} = \mathcal{L}_{EC}+\lambda_{1}\mathcal{L}_{CC}+\lambda_{2}\mathcal{L}_{RC}+\lambda_{3}\mathcal{L}_{AL}.
	\end{aligned}
\end{equation}

\subsection{\textbf{Mixture of Experts}}
Fig.\ref{fig:4} shows the process of multi-expert knowledge fusion and decision-making. In this process, the visual-text matching is used to calculate the similarity between visual and textual features to achieve identity prediction. Meanwhile, the mixture of experts (MoE) module aims to integrate visual-textual features with profound insights from multiple experts, achieving deep fusion of all complementary features. MoE consists of two modules of the visual-textual fusion (VTF) and the expert gating network $G(\bigcdot)$. The VTF module consists of three parts of self-attention (SA), cross-attention (CA) and ReID-Head projector. It aims to compute the similarity and integrate complementarity between visual and textual features. The expert gating network $G(\bigcdot)$ can be viewed as a mapping function $\mathbb{R}^{b \times d}\rightarrow \mathbb{R}^{b \times k}$, where $b$ denotes the number of instance number of each batch and $d$ represents the feature dimension. For the visual features $f_{n}^k\in\mathbb{R}^{b \times d}$ and text features $T_{N}^k\in\mathbb{R}^{b \times d}$, MoE can be used to calculate the weighted score of the identity predictions of different experts for both features. The formulation of the weighted sum $Z_{t}$ of the mixture of experts is calculated as below:
\begin{equation}\label{formula15}
	\begin{aligned}
		Z_{t}=\sum_{k=1}^{K}G_{k}(F_{C})\bigcdot m_{k}(F_{C}),
	\end{aligned}
\end{equation}
where the ReID-Head projector $m_{k}(\bigcdot)$ is composed of $K$ parallel projections for each expert, the attention-based multi-expert features $F_{C} \in \mathbb{R}^{b \times d}$ is obtained by entering visual and textual features together into SA and CA. As mentioned in reference \cite{xue2022one}, MoE can be treated as specialists and assign knowledge to the student. In MiKeCoCo, we use knowledge distillation (KD) to guide the training of ID classifier through $\mathcal{L}_{dis}$. The target of the loss function is to minimize the Kullback-Leibler (KL) divergence between the output of teacher and student for the training of image encoder of CLIP. The distillation loss of this module is as below:
\begin{equation}\label{formula15}
	\begin{aligned}
		\mathcal{L}_{dis} = \mathcal{L}_{KL}\big(\sigma(Z_{s}), \sigma(Z_{t})\big)=\sum_{}Z_{s}log\big(\frac{Z_{s}}{Z_{t}}\big),
	\end{aligned}
\end{equation}
where $Z_{s}$ and $Z_{t}$ denote the the logits of student and teacher model, respectively. $\sigma(\cdot)$ is the softmax function. Consequently, multi-expert knowledge distillation can significantly enhance the student model performance beyond what has been achieved through only supervised learning.

\subsection{Model training and inference}

In the first training stage, the image encoder and text encoder of CLIP are not optimized, text representations and image representations extracted by adversarial learning are aligned through visual-text contrastive learning which has been shown to be an effective objective for improving vision and language understanding.
We fix the parameters of the visual encoder and textual encoder of CLIP and then only update the parameters of the MEKA module and the learnable prompt set. The overall losses of this stage is calculated as:
\begin{equation}\label{formula18}
	\begin{aligned}
\mathcal{L}_{stage1}=\mathcal{L}_{MEKA}+\mathcal{L}_{v2t}+\mathcal{L}_{t2v},
	\end{aligned}
\end{equation}
where $\mathcal{L}_{v2t}$ and $\mathcal{L}_{t2v}$ are visual-to-text contrastive loss and text-to-visual contrastive loss, the settings of them follow that in the first training stage of CLIP-ReID. 

In the second training stage, we have obtained the multi-expert visual from MEKA and the learned prompt features. To guarantee that the image encoder captures more robust and comprehensive visual features, style perturbation images are utilized in this stage. In addition, we introduce the MoE module which utilizes the gating network and the visual-text fusion block to control the contribution of each expert. This MoE module acts as a teacher model and efficiently guides the student model for vehicle ReID. The ID loss $\mathcal{L}_{ID}$ can be computed as:
\begin{equation}\label{formula19}
	\begin{aligned}
		\mathcal{L}_{ID} = \sum_{n=1}^{N}-q_{n}\bigcdot log(p_{n}),
	\end{aligned}
\end{equation}
where $q_{n}$ denotes the target distribution and $p_{n}$ indicates the prediction logits of $n$-$th$ ID from ID classifier.

Furthermore, building on the cross-entropy $\mathcal{L}_{v2tce}$ in CLIP-ReID, we combine the prediction results of multiple experts, each calculating the similarity between visual and textual features using $\mathcal{L}_{v2tce}$. Then, $\mathcal{L}_{dis}$, $\mathcal{L}_{ID}$ and $\mathcal{L}_{v2tce}$ are employed to form the total losses in the second stage. Further details on the settings of $\mathcal{L}_{v2t}$, $\mathcal{L}_{t2v}$, and $\mathcal{L}_{v2tce}$ are presented in the supplementary materials. We use the weighted factors $\alpha_{1}$ and $\alpha_{2}$ for $\mathcal{L}_{ID}$ and $\mathcal{L}_{v2tce}$. The overall losses of this stage is calculated as:
\begin{equation}\label{formula21}
	\begin{aligned}
		\mathcal{L}_{stage2}=\alpha_{1}\mathcal{L}_{ID}+\alpha_{2}\mathcal{L}_{v2tce}+\mathcal{L}_{dis}.
	\end{aligned}
\end{equation}

During inference, only the original images are fed into the image encoder, which is trained in the second stage, to extract visual feature vectors for the retrieval results.
\section{Experiment}
\subsection{Datasets and Evaluation Protocol}
We conduct experiments on four datasets: VeRi--776 \cite{r19}, VehicleX \cite{r46}, VehicleID \cite{r47} and Opri \cite{r71}. The statistical information for these datasets is summarized in Table~\ref{vehicle-dataset}. This paper follows four domain generalization experiment settings, which include VeRi--776$\to$Opri, VehicleX$\to$Opri, VeRi--776$\to$VehicleID, and VehicleX$\to$VehicleID. In the VeRi--776$\to$VehicleID setting, the three evaluation sizes (1600, 2400, 3200) of the VehicleID dataset are denoted as \textbf{Protocol-1}, \textbf{Protocol-2}, and \textbf{Protocol-3}. Similarly, in the VehicleX$\to$VehicleID setting, they are denoted as \textbf{Protocol-4}, \textbf{Protocol-5}, and \textbf{Protocol-6}. According to the usual practice, the Cumulative Matching Characteristics (CMC) at Rank-1 and the mean Average Precision (mAP) are used as performance evaluation metrics.

\begin{table}[t!]
\caption{Statistical analysis of the vehicle datasets.}
    \scriptsize
    \centering
    \resizebox{\columnwidth}{!}
    {
    \begin{tabular}{c|cccc}\hline
         \textbf{Vehicle Datasets}    &  \textbf{IDs} &  \textbf{images} & \textbf{cameras} & \textbf{scale}    \\ \hline
         VeRi--776 \cite{r19} & 776 & 49,360 & 20 & medium \\
         VehicleX \cite{r46}  & 1,362 & 192,150 & 11 & large\\
         VehicleID \cite{r47} & 26,267 & 221,763 & 12 & large \\
         Opri \cite{r71}    & 17,835 & 130,994 & -- & large  \\
         \hline
    \end{tabular}
    }
    \label{vehicle-dataset}
\end{table}

\begin{table}[t!]
 \caption{
    The comparison results (\%) of the state-of-the-art methods on VeRi--776$\to$Opri and VehicleX$\to$Opri settings. The best and second best results are marked in \textbf{bold} and \underline{underline}, respectively.
    }
    \centering
    \resizebox{\columnwidth}{!}
    {
    \begin{tabular}{c|c|cc|cc}\hline
         \multirow{2}{*}{\textbf{Methods}} &\multirow{2}{*}{\textbf{Venue}}   &    \multicolumn{2}{c|}{\textbf{VeRi--776$\to$Opri}}  &   \multicolumn{2}{c}{\textbf{VehicleX$\to$Opri}}            \\
         \cline{3-6}
             &    & \textbf{mAP}   & \textbf{Rank-1} & \textbf{mAP}   & \textbf{Rank-1}
              \\ \hline
         IS-GAN
                            & NIPS'19
                     &   26.4    & 37.8 &   25.8    & 35.3
                     \\
         SNR
                            & CVPR'20
                     &   21.2    & 31.5 &   21.2    & 31.6
                          \\
         CAL
                           & ICCV'21
                       & 17.1    &26.3 & 7.0    &11.7
                                                  \\
         MetaBIN
                            & CVPR'21
                           & 22.0 & 32.5 & 21.7 & 31.0
                          \\
         MixStyle
                            &ICLR'21
                           & 26.4 & 36.2 & 20.6 & 29.4
                            \\
         OSNet-AIN
                            &TPAMI'21
                        & 26.1   & 36.5 & 13.4   & 20.4
                        \\
         TransMatcher
                           &NIPS'21
                           & 21.9 & 34.1 & 15.6 & 24.4
                           \\
         TransMatcher-GS
                            &NIPS'21
                           & 34.7 & 42.8 & 28.9 & 36.7
                           \\
        QAConv-GS
                            &CVPR'22
                           & 35.2  & 44.6 &\textbf{34.1} & \textbf{43.0}
                           \\
        JIFD
                             & T--ITS'23
                           & 25.3  & 35.0  & 23.0  & 32.8
                           \\
        MSI-Net
                           & CVPR'23
                        & 25.2   &35.9 & 18.0   &26.7
                        \\
        PAT
                           & ICCV'23
                       & 28.5    &38.7  & 26.3    &36.0
                           \\
        CLIP-ReID
                        & AAAI'23
                        & 21.2   & 30.8  & 21.7    & 30.5
                          \\
         SC
                       & MM'23
                       &  32.2   & 42.4    &  29.7  & 39.8
                       \\
         IAD
                       & NN'24
                       &  20.9  &  30.9   &  20.6  & 29.7
                          \\
        MiKeCoCo! \textbf{(Ours)}  &  This Venue & \underline{39.1} & \textbf{50.1} & 31.2 & 41.1  \\
        MiKeCoCo \textbf{(Ours)}  &  This Venue & \textbf{39.3} & \underline{49.6} & \underline{32.8} & \underline{41.5} \\
        \hline
    \end{tabular}}
    \label{SOTA1}
\end{table}

\begin{table*}[t!]
\caption{The comparison results (\%) of the state-of-the-art methods on VeRi--776$\to$VehicleID and VehicleX$\to$VehicleID settings. The best and second best results are marked in \textbf{bold} and \underline{underline}, respectively.}
    \centering
    \resizebox{\linewidth}{!}{
    \begin{tabular}{c|c|cc|cc|cc|cc|cc|cc}\hline
         \multirow{2}{*}{\textbf{Methods}}&\multirow{2}{*}{\textbf{Venue}}& \multicolumn{2}{c|}{\textbf{Protocol-1}}  & \multicolumn{2}{c|}{\textbf{Protocol-2}}
          & \multicolumn{2}{c|}{\textbf{Protocol-3}} & \multicolumn{2}{c|}{\textbf{Protocol-4}}  & \multicolumn{2}{c|}{\textbf{Protocol-5}}
          & \multicolumn{2}{c}{\textbf{Protocol-6}}
         \\ \cline{3-14}
             &     & \textbf{mAP}  & \textbf{Rank-1}
                   & \textbf{mAP}  & \textbf{Rank-1}
                   & \textbf{mAP}  & \textbf{Rank-1}
                   & \textbf{mAP}  & \textbf{Rank-1}
                   & \textbf{mAP}  & \textbf{Rank-1}
                   & \textbf{mAP}  & \textbf{Rank-1}
                   \\ \hline
         Dual-branch
                            & WACV'20
                           & 51.6 & 47.3
                           & -- & --
                           & 45.3 & 41.2
                           & -- & --
                           & -- & --
                           & -- & --
                      \\
         Dual-branch+CaNE
                            & WACV'20
                           & 53.1 & 48.7
                           & -- & --
                           & 46.3 & 42.1
                           & -- & --
                           & -- & --
                           & -- & --
                     \\
         SNR
                            & CVPR'20
                           & 35.1 & 28.8
                           & 30.6 & 24.5
                           & 29.4 & 23.6
                           & 33.4 & 27.2
                           & 31.1 & 25.1
                           & 29.2 & 23.3
                          \\
         CAL
                            & ICCV'21
                          & 49.8 & 40.5
                          & 44.6 & 35.4
                          & 42.9 & 34.3
                          & 22.2 & 15.8
                           & 19.4 & 13.6
                           & 17.9 & 12.6
                          \\
         MixStyle
                            &ICLR'21
                           & 55.9 & 47.3
                           & 52.1 & 43.3
                           & 48.9  & 40.3
                           & 41.4 & 33.3
                           & 38.0 & 30.4
                           & 35.6  & 28.1
                           \\
         TransMatcher
                            &NIPS'21
                          &59.2 & 51.6
                          &56.3 & 48.4
                          &53.0 & 45.4
                          & 53.5 & 45.9
                           & 49.8 & 41.8
                           & 46.7 & 39.0
                          \\
         TransMatcher-GS
                            &NIPS'21
                           & \textbf{61.5} & 52.9
                           & \textbf{57.5} & 48.5
                           & 53.3 & 44.4
                           &  49.2 & 41.0
                           &  45.2 & 36.7
                           &  36.7 & 33.3
                          \\
         QAConv-GS
                            &CVPR'22
                           & 59.2 & 52.0
                           & 56.2 & 49.0
                           & 52.6 & 45.2
                           & 49.5 & 42.2
                           & 45.6 & 38.1
                           & 42.8 & 35.6
                           \\
        JIFD
                             & T--ITS'23
                           & 44.4  & 36.8
                           & 40.0  & 32.4
                           & 37.7  & 30.7
                           & 35.2  & 27.8
                           & 31.5  & 24.8
                           & 29.4  & 23.3
                           \\
         MSI-Net
                            & CVPR'23
                          & 56.7 & 47.9
                          & 53.5 & 44.2
                          & 49.5 & 41.3
                           & 38.4 & 30.5
                           & 34.9 & 27.4
                           & 34.9 & 25.3
                          \\
         PAT
                          & ICCV'23
                           & 53.1    &44.5
                           & 49.5    &40.8
                           & 46.3    &38.1
                            & 46.3    &38.2
                            & 42.1    &34.3
                            & 39.7    &32.3
                           \\
         CLIP-ReID
                           & AAAI'23
                           & 52.3 & 42.2
                           & 47.9 & 38.0
                           & 44.3 & 34.7
                           & 43.8 & 34.8
                           & 39.2 & 30.5
                           & 36.8 & 28.7
                           \\
         SC
                       & MM'23
                           & 48.4 & 41.1
                           & 44.8 & 37.6
                           & 42.6 & 35.3
                           & 48.2 & 41.0
                           & 44.4 & 36.6
                           & 42.2 & 34.9
                           \\
         IAD
                            & NN'24
                           & 54.3 & 44.2
                           & 50.9 & 41.3
                           & 46.0 & 36.7
                           & 37.7 & 29.7
                           & 33.7 & 26.0
                           & 31.3 & 24.4
                            \\
         MiKeCoCo! \textbf{(Ours)}  &  This Venue & 60.3 & \underline{54.1}
                                                  &56.4  &\underline{49.5}
                                                  &\underline{54.0} &\underline{47.0}
                                                & \textbf{55.4} & \textbf{48.0}
                                                & \underline{50.5} & \underline{42.5}
                                                & \underline{48.1} & \underline{40.5}\\
         MiKeCoCo \textbf{(Ours)}  &  This Venue  &\underline{61.1}  &\textbf{54.4}
                                                  &\underline{57.3}  &\textbf{49.8}
                                                  &\textbf{55.0}
                                                  &\textbf{48.0}
                                                   & \underline{54.9} & \underline{47.2}
                                                 & \textbf{51.9} & \textbf{43.8}
                                                 & \textbf{48.2} & \textbf{40.5}
                     \\ \hline
    \end{tabular}}
    \label{SOTA3}
\end{table*}

\subsection{Implementation Details}
We adopt the ViT-based model as the visual encoder for CLIP, keeping the other experimental settings consistent with CLIP-ReID \cite{r18}. In the expert gating network, we concatenate the fused features with the text features and use a fully-connected layer to obtain the score of each expert. The Adam optimizer \cite{kingma2014adam} is used with a learning rate initialized at $3.5\times 10^{-4}$, which decays according to a cosine schedule. We use the $P \times M$ sampler to construct the training mini-batch, where $P$ denotes the number of identities and $M$ indicates the number of images per identity. The model is optimized using a warm-up learning rate that increases from $5.0\times 10^{-7}$ to $5.0\times 10^{-6}$. To balance the different loss terms, the coefficients $\lambda_{1}$, $\lambda_{2}$, and $\lambda_{3}$ in $\mathcal{L}_{stage1}$ are set to 0.1, 10.0, and 0.2, respectively. The hyperparameters $\alpha_{1}$ and $\alpha_{2}$ in $\mathcal{L}_{stage2}$ are set to 0.25 and 1.8. We also apply random cropping, horizontal flipping, and random rotation to the input images. MiKeCoCo is trained on a single 3090Ti GPU during both the first and second stages. MiKeCoCo! and MiKeCoCo denote the number of experts, 2 and 3 respectively. All evaluations strictly follow the single-query evaluation protocol.

\subsection{Comparison with State-of-the-Art Methods}
To demonstrate the advanced performance of our methods, we compare our method with several recent state-of-the-art (SOTA) domain generalization methods, including IS-GAN~\cite{r96}, Dual-branch~\cite{r66}, Dual-branch+CaNE~\cite{r66}, SNR~\cite{r61}, CAL~\cite{r62}, MetaBIN~\cite{r23}, Mixstyle~\cite{r63}, OSNet-AIN~\cite{r64}, Transmatcher~\cite{r7}, Transmatcher-GS~\cite{r7}, JIFD~\cite{r71}, MSI-Net~\cite{r65}, QAConv-GS~\cite{r8}, PAT~\cite{r70}, CLIP-ReID~\cite{r18}, SC~\cite{s2}, and IAD~\cite{s1}. The specific experimental results are shown in Tables~\ref{SOTA1} and~\ref{SOTA3}. The `-GS' suffix indicates that the graph sampling strategy~\cite{r8} is used. All other CLIP-ReID results presented in this paper adhere to the default parameters from the official codebase. MiKeCoCo method demonstrates its outstanding performance in experimental comparisons.

\subsection{Ablation Studies and Analysis}

\begin{table}[t!]
\caption{
    The ablation experiments on different components. \textbf{MEKA} means that the model is only optimized with visual-text matching. \textbf{KD} means the Knowledge Distillation.
    }
    \label{ablation1}
    \scriptsize  
    \centering
    \resizebox{\columnwidth}{!}{
    \begin{tabular}{c|cccc|cc|cc}\hline
        \multirow{2}{*}{}
        &  \multicolumn{4}{c}{\textbf{Different Components}}&  \multicolumn{2}{|c|}{\textbf{VeRi--776$\to$Opri}} & \multicolumn{2}{c}{\textbf{VehicleX$\to$Opri}}\\ \cline{2-9}
          &  $\textbf{MEKA}$ & $\textbf{MoE}$  & $\textbf{VTF}$ & $\textbf{KD}$  & \textbf{mAP} & \textbf{Rank-1} & \textbf{mAP}  & \textbf{Rank-1} \\
        \hline
        A & \checkmark &  &  &      &  37.6  & 47.4   & 31.5  & 40.8   \\
        B &  \checkmark & \checkmark  & &  &  37.7  &  47.8  & 32.1  & 40.9 \\
        C &  \checkmark & \checkmark  & \checkmark   &   & 37.3   & 47.8   & 32.0  & 41.4   \\
        D &  \checkmark & \checkmark  & &\checkmark      & 38.8   & 48.7   & 31.8  & 41.0   \\
        E &  \checkmark & \checkmark  & \checkmark  & \checkmark  &  \textbf{39.3}  & \textbf{49.6}  & \textbf{32.8}   & \textbf{41.5} \\
        \hline
    \end{tabular}
}
\end{table}

\begin{table}[ht!]
    \caption{
    The effects of inputs from different image components on TransMatcher-GS, CLIP-ReID and MiKeCoCo models.
    }
    \label{ablation2}
    \centering
    \scriptsize
    \resizebox{\columnwidth}{!}{
    \begin{tabular}{c|cc|cc|cc}\hline 
    \multirow{2}{*}{\textbf{Methods}} & 
    \multirow{2}{*}{\textbf{Training}} &
    \multirow{2}{*}{\textbf{Test}}  &
    \multicolumn{2}{c|}{\textbf{VeRi--776$\to$Opri}} & \multicolumn{2}{c}{\textbf{VehicleX$\to$Opri}}\\ \cline{4-7}
      & & &
      \textbf{mAP} & \textbf{Rank-1} & \textbf{mAP}  & \textbf{Rank-1} \\ \hline
 \multirow{4}{*}{TransMatcher-GS} & $ X_{s}$ & $X_{t}$  & 34.7   & 42.8  & 28.9    & 36.7   \\ 
       & DII & $X_{t}$  & 37.5 & 46.7 & 31.5  &  40.2  \\ 
       & DII & DII  & 38.4 & 48.0 & \textbf{32.0} & \textbf{40.7} \\
       & SPI & $X_{t}$  & \textbf{39.6} & \textbf{49.0} & 29.8 & 39.2
       \\ \hline\hline

        \multirow{2}{*}{\textbf{Methods}} & \multicolumn{2}{c|}{\textbf{Training Modes}}   &  \multicolumn{2}{c|}{\textbf{VeRi--776$\to$Opri}} & \multicolumn{2}{c}{\textbf{VehicleX$\to$Opri}}\\ \cline{2-7}
      &  $\textbf{Stage-1}$ & $\textbf{Stage-2}$  & \textbf{mAP} & \textbf{Rank-1} & \textbf{mAP}  & \textbf{Rank-1} \\ \hline
       \multirow{3}{*}{CLIP-ReID} & $ X_{s}$ & $X_{s}$  & 21.2   & 30.8  & 21.7    & 30.5   \\ 
       & SPI & SPI  & 22.9 & 32.3 & 25.4  &  33.9  \\ 
       & DII & SPI  & \textbf{23.8} & \textbf{32.9} & \textbf{26.2} &  \textbf{34.8}   \\ \hline

      \multirow{3}{*}{MiKeCoCo!} & $X_{s}$ & $X_{s}$  & 34.2 & 44.3 & 29.2 & 38.9  \\ 
       & SPI & SPI & 36.2  & 46.7 & 30.1 & 39.6   \\ 
      & DII & SPI & \textbf{39.1} & \textbf{50.1} & \textbf{31.2} & \textbf{41.1}   \\ \hline

      \multirow{3}{*}{MiKeCoCo} & $X_{s}$ & $X_{s}$  & 34.3 & 45.4 & 30.0 & 39.7   \\ 
       & SPI & SPI  & 37.7 & 48.3 & 30.8 & 40.4      \\ 
       & DII & SPI & \textbf{39.3} & \textbf{49.6} & \textbf{32.8} & \textbf{41.5}    \\ 
        \hline
    \end{tabular}
}
\end{table}

\begin{figure}[t!]
	\centering
	\includegraphics[width=2.8in,height=1.85in]{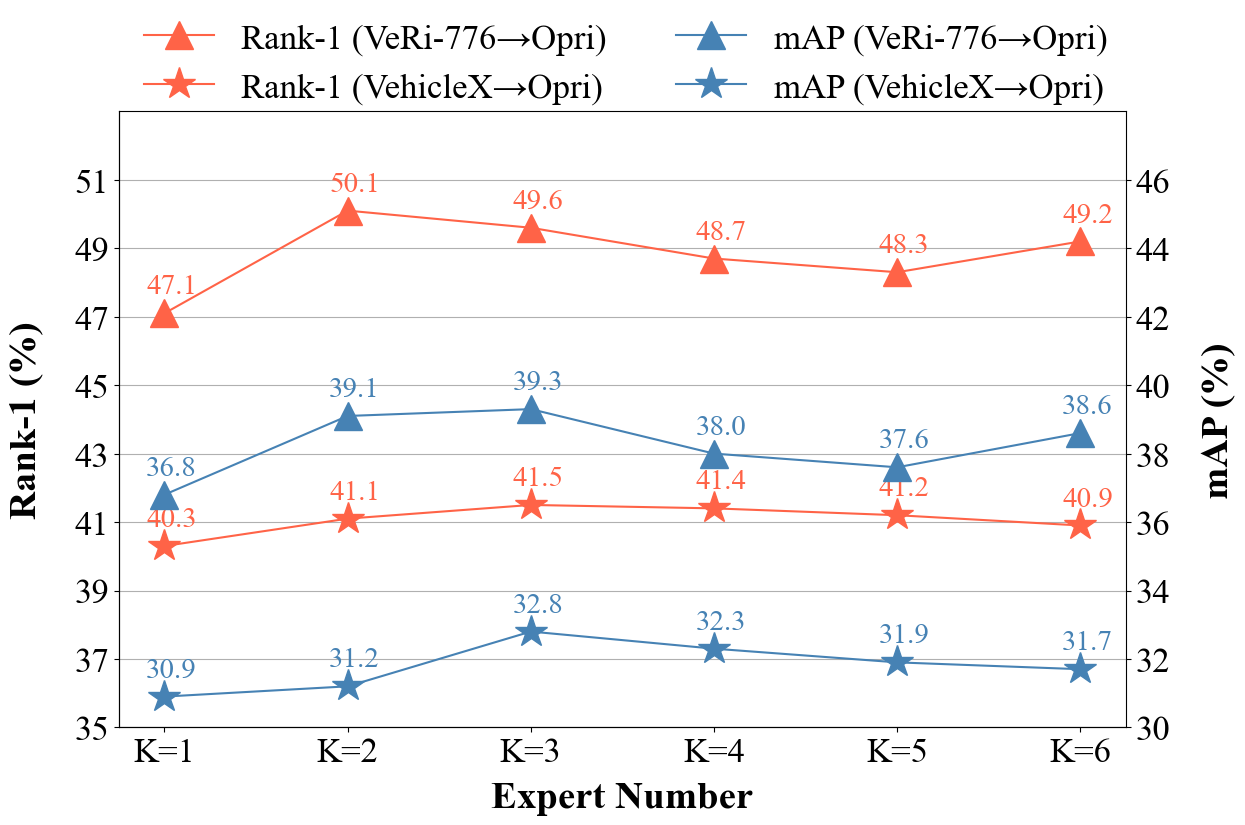}
	\caption{Parameter sensitivity analysis of the number of experts on VeRi--776$\to$Opri and VehicleX$\to$Opri settings.}
	\label{ablation3}
\end{figure}

In the ablation experiment, we firstly conduct comparative experiments on different components of our model and number of experts. Specifically, by controlling different components of the model one by one, we observe that each component positively contributes to the model's final performance, as shown in Table \ref{ablation1}. When all components work together, the model achieves the optimal performance state. On the other hand, the experiments on the number of experts reveal the potential of the multi-expert system, as shown in Fig.\ref{ablation3}. These results show that MiKeCoCo achieves the best recognition performance when the number of experts is 2 or 3. This result not only demonstrates that a multi-expert system can enhance model performance through teamwork but also suggests the need to balance the number of experts with unique perspective in practical applications to achieve optimal feature allocation.

To verify the validity of the STREAM scheme, different methods are trained and tested using DII and SPI respectively, as shown in Table \ref{ablation2}. When DII and SPI are fed into the first and second stages of MiKeCoCo, our method achieves the best recognition performance. The results of other ablation comparison experiments further demonstrate the advantages of domain-dependent redundancy elimination and style perturbation.

\section{Conclusion}
This paper proposes a dual-drive framework to tackle domain generalization issues in vehicle ReID. It combines a data-driven approach focused on eliminating image redundancy with a model-driven strategy that leverages multi-expert perspectives obtained by adversarial learning. By eliminating domain-related redundancies in the images, it enables the model to focus more closely on the detailed complementary features, which are crucial for distinguishing fine-grained vehicles. The comprehensive representation integrated with complementary features is crucial in many practical application scenarios. The experimental results validate the effectiveness of our proposed method.

{
    \small
    \bibliographystyle{ieeenat_fullname}
    \bibliography{main}
}


\end{document}